\pdfoutput=1
\PassOptionsToPackage{table}{xcolor}
\documentclass[letterpaper]{article}
\usepackage[preprint]{aaai2027}
\usepackage[hyphens]{url}
\usepackage{graphicx}
\usepackage{natbib}
\usepackage{caption}
\usepackage{multirow}
\usepackage{makecell}
\usepackage{booktabs,tabularx,amssymb}
\usepackage{xcolor}
\usepackage{amsmath}
\DeclareCaptionStyle{ruled}{labelfont=normalfont,labelsep=colon,strut=off}
\newcolumntype{C}{>{\centering\arraybackslash\hspace{0pt}}X}

\title{UniCSI\\Towards a Universal Wi-Fi CSI Encoder for Ubiquitous Human Sensing}
\author {
    Daniel Eckhoff \textsuperscript{\rm 1},
    Hua Kang \textsuperscript{\rm 1},
    Zhitang Chen \textsuperscript{\rm 1},
    Jie Chuai \textsuperscript{\rm 1}
}
\affiliations {
    \textsuperscript{\rm 1}Huawei Noah's Ark Lab\\
    \{eckhoff.daniel1,kang.hua1,chenzhitang2,chuaijie\}@huawei.com
}

\begin{document}

\maketitle

\begin{abstract}
Wi-Fi sensing promises to turn the everyday wireless signals that already surround us into ubiquitous sensors for human sensing. However, a fundamental obstacle is that CSI is acquired under diverse device-specific configurations, including different subcarrier counts, bandwidths, and carrier bands. Consequently, the resulting CSI tensors vary in both spectral resolution and tensor shape, making heterogeneous modeling challenging. Standard architectures struggle with such heterogeneity, forcing lossy pre-processing which compromises the underlying signal. To bridge this gap, we present UniCSI, a unified foundation architecture that directly operates on heterogeneous CSI while preserving the integrity of the native waveform. UniCSI hinges on two core innovations: (1) a physics-informed RF tokenizer that encodes each frequency channel based on its fractional position within the physical spectrum rather than rigid array indices. It preserves intrinsic spectral coherence and enables seamless, frequency resolution-agnostic processing across arbitrary sensing configurations. (2) a spectral aggregator that distills variable-length channel sequences into a fixed-size spectral signature, effectively decoupling the feature dimensionality from the physical subcarrier spacing. Extensive evaluations on a large-scale corpus of 25 heterogeneous public datasets, spanning 14 to 2048 subcarriers, 20 to 160~MHz bandwidth, and the 2.4 and 5~GHz bands, demonstrate that native heterogeneous ingestion substantially improves cross-domain transfer under both supervised and self-supervised training schemes, particularly in regimes where fixed-grid architectures fail to generalize.
\end{abstract}

\section{Introduction}
Wireless sensing has emerged as a promising paradigm for ubiquitous human-centric perception, with commodity Wi-Fi devices supporting applications from activity recognition and gesture interaction to health monitoring and indoor localization~\cite{wang2014a,zhang2021a,abdelnasser2015,adib2014}. Compared with vision-based sensing, it preserves privacy, operates under poor lighting and occlusion, and leverages existing infrastructure without wearables or dedicated sensors. Among wireless signals, Channel State Information (CSI) has become the dominant sensing modality due to its fine-grained characterization of wireless propagation and its sensitivity to environmental dynamics.

Despite the remarkable progress of Wi-Fi sensing, existing solutions suffer from poor cross-domain generalization because CSI measurements are inherently sensitive to environments, devices, and wireless configurations. The resulting distribution shifts significantly limit model transferability across deployment scenarios.
Large-scale pretraining offers a promising solution, learning transferable representations from diverse datasets. Unlike vision datasets, however, which share a common representation format, Wi-Fi sensing datasets are inherently heterogeneous, differing in signal dimensions and acquisition configuration such as subcarrier count, bandwidth, carrier frequency, number of devices, and antenna configuration. These differences are not only observed across devices and datasets, but may also vary over time for the same device as communication parameters are dynamically reconfigured.

Existing solutions~\cite{jiang2025,zhu2026,luo2026,kim2026} handle this heterogeneity through ad hoc preprocessing, such as zero-padding, subcarrier resampling, or per-device backbones, each of which wastes capacity, distorts the signal, or precludes knowledge sharing. None yields a single model that exploits heterogeneous datasets within a unified framework.

To address these limitations, we introduce UniCSI, a carrier-frequency- and bandwidth-agnostic foundation model for Wi-Fi sensing, built on two components. The first is a per-subcarrier tokenizer that accepts any subcarrier count without padding or resampling, encoding each subcarrier by its fractional position within the observed band rather than by a fixed array index, so that the representation depends on where a subcarrier sits in the band rather than on how finely the band is sampled. Because each antenna link is embedded separately by this same tokenizer, the scheme extends to any MIMO configuration without assuming a fixed antenna layout. The second is a cross-attention module that compresses variable-length spectra into a fixed latent representation, independent of input length and sampling density. Together, these enable a single model to ingest CSI of any shape natively.

In particular, this work makes the following contributions. (1) We propose, to the best of our knowledge, the first CSI encoder that natively ingests measurements of arbitrary subcarrier count, bandwidth, and carrier band without padding or resampling, and that extends to varying MIMO configurations through per-link encoding. (2) We show that JEPA pretraining on a large heterogeneous CSI corpus yields transferable representations, and characterize how much this pretraining benefits from heterogeneity across datasets, bands, and bandwidths. (3) We show that on difficult cross-domain benchmarks, fixed-grid architectures collapse toward chance while UniCSI transfers,
establishing native heterogeneous ingestion as a prerequisite for
cross-domain generalization rather than a convenience.

\subsection{Related Work}
Wi-Fi sensing has evolved from CNN- and RNN-based models toward Transformer architectures that better capture long-range dependencies \cite{sensefi,zhang2021a,witransformer,metafipp,wiflexformer}. These models, however, are typically designed for a single task under homogeneous conditions, and transfer poorly once devices, environments, or configurations shift. This fragility has motivated a line of Wi-Fi sensing foundation models, which share a single backbone across tasks but differ in how, and how much, they confront the heterogeneity of CSI.

\citet{jiang2025} harmonize datasets into a common format before pretraining, converting every sample to amplitude, forcing a canonical single-transmitter three-receive-antenna layout, and resampling the temporal and subcarrier axes to fixed lengths to yield a fixed $600 \times 90$ input. This resample-to-a-fixed-grid strategy is precisely what we seek to avoid, since interpolating the subcarrier axis discards the physical spacing of the measurement and normalizes away the heterogeneity a foundation model should exploit.

WiSenseNet~\citep{lyons2025} and AM-FM~\citep{zhu2026} likewise resolve heterogeneity before or beneath the model rather than in it: the former mixes bandwidths in training but relies on subcarrier-selection preprocessing and evaluates each task within a single benchmark, while the latter compresses the frequency axis with cross-attention yet does so on top of a fixed padded grid with subcarrier and antenna dimensions collapsed into one axis.

CSI-JEPA \citep{luo2026} is the first application of joint-embedding predictive pretraining to Wi-Fi CSI, and contributes a channel variation-aware masking strategy that targets high-variation regions rather than masking uniformly. Like AM-FM, however, it tokenizes on a fixed $P_K \times P_T$ patch grid with index-based positional embeddings, so heterogeneity in subcarrier count and bandwidth must be resolved before tokenization; its evaluation likewise stays within a single benchmark.

Closest to our own motivation, WiFi-JEPA \citep{kim2026} argues against flattening CSI into a 2D grid, since patch embedding then crosses link and amplitude-phase boundaries, and instead keeps the $(C, T, L)$ structure intact. Its central contribution is link masking, which predicts one antenna-link view from the others to exploit cross-link redundancy. WiFi-JEPA, however, is specialized to a single fixed acquisition configuration and tokenizes with index-based positional embeddings, so unlike UniCSI it neither ingests varying subcarrier counts and bandwidths nor encodes subcarriers by physical band position.

In contrast, UniCSI treats variation in subcarrier count, bandwidth, and carrier band not as something to be normalized away before the model, as the preceding methods do, but as something the architecture ingests natively, encoding each subcarrier by its fractional position within the observed band so that a single model spans differing shapes rather than one fixed configuration.Since these methods share fixed-grid tokenization as their ingestion substrate, our experiments compare against controlled instantiations of that substrate under matched data and training conditions.

\section{Method}
\paragraph{Preliminaries of Wi-Fi CSI}
Wi-Fi operates in the 2.4 and 5~GHz bands using orthogonal frequency-division multiplexing (OFDM), which divides the channel into a set of narrow, evenly spaced frequency components called \emph{subcarriers}. Because each subcarrier probes the channel at a slightly different frequency, a CSI measurement over an observation window is a complex-valued tensor
\begin{equation}
\mathbf{H} \in \mathbb{C}^{T \times N_{\mathrm{tx}} \times N_{\mathrm{rx}} \times K},
\label{eq:csi_tensor}
\end{equation}
whose element $\mathbf{H}_{t,i,j,k}=a e^{\mathrm{j}\phi}$ records the amplitude $a$ and phase $\phi$ of the channel between transmit antenna $i$ and receive antenna $j$ at time step $t$ and subcarrier $k$. Here $T$ is the number of packets, $N_{\mathrm{tx}}$ and $N_{\mathrm{rx}}$ the transmit and receive antenna counts, and $K$ the number of subcarriers. Crucially, both the MIMO configuration $(N_{\mathrm{tx}}, N_{\mathrm{rx}})$ and the subcarrier count $K$ depend on the underlying hardware and bandwidth, so CSI arrives with heterogeneous dimensions across devices, and an effective foundation model must accommodate them without fixed input shapes or device-specific architectures.

Throughout, we operate on the amplitude $a$ and discard the phase $\phi$. This choice follows from both the data landscape and our design goals. Most public CSI datasets release amplitude only, making it the common denominator across corpora and the feature most used in prior work \citep{jiang2025,zhu2026}. Raw phase, moreover, is corrupted by carrier-frequency, sampling-clock, and packet-detection offsets, and its sanitization is chipset-specific, precisely the configuration-dependent preprocessing that UniCSI is designed to avoid.
\paragraph{Overview}
\begin{figure*}
    \centering
    \includegraphics{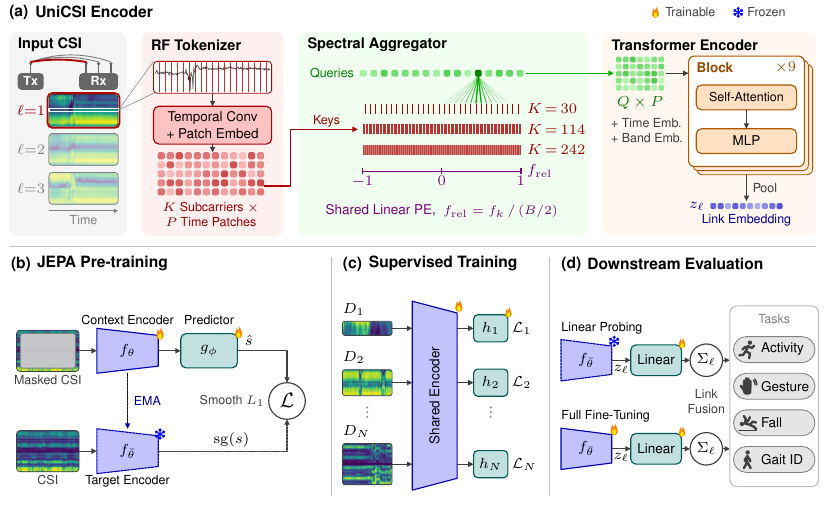}
    \caption{Overview of UniCSI. \textbf{(Top)} Per-link encoder: a temporal CNN and time-patch projection tokenize per-subcarrier CSI; learned queries with band-relative positional encoding cross-attend into the subcarrier tokens, compressing the variable-length spectrum into a fixed $Q \times P \times d$ latent for a Transformer encoder. \textbf{(Bottom)} Training regimes: JEPA pretraining, supervised multi-task training, and downstream probing or fine-tuning.}
    \label{fig:framework}
\end{figure*}

Given a CSI measurement, our goal is to train a unified encoder that produces a fixed-dimensional representation regardless of the underlying Wi-Fi hardware and communication parameters. In particular, the encoder should naturally accommodate heterogeneous CSI measurements with varying numbers of subcarriers and antenna configurations, without requiring zero-padding, resampling, or device-specific backbones.

We describe \textbf{UniCSI}, a heterogeneous CSI encoder composed of three key components as shown in Figure~\ref{fig:framework}. First, each transmit--receive antenna link is processed independently by a physics-informed RF tokenizer to capture local temporal dynamics for every subcarrier. Next, a spectral aggregator explicitly models dependencies across subcarriers while remaining agnostic to the number of input subcarriers. Finally, the aggregated tokens are fed into a Transformer encoder to produce a fixed-dimensional representation for downstream Wi-Fi sensing tasks. This architecture can be trained under either supervised multi-task learning or self-supervised JEPA pretraining without modification.

\paragraph{Physics-Informed RF Tokenizer}
UniCSI processes each transmit--receive antenna link independently and tokenizes each subcarrier separately, rather than partitioning the entire $K\times T$ CSI matrix into two-dimensional patches. This design is motivated by the asymmetric semantics of CSI: the temporal axis captures channel evolution over time, whereas the subcarrier axis represents frequency-domain responses. Moreover, subcarrier-wise tokenization naturally accommodates heterogeneous CSI with varying numbers of subcarriers, avoiding fixed patch layouts tied to specific hardware configurations. Specifically, a lightweight 1D CNN is first applied to each subcarrier independently to capture local temporal patterns. The resulting features are then divided into $P=\lfloor T/\tau \rfloor$ non-overlapping windows of length $\tau$, each of which is linearly projected into a $d$-dimensional RF token, yielding
$
\mathbf{Z}\in\mathbb{R}^{K\times P\times d}.
$
Since tokenization is performed independently for each subcarrier, $K$ only determines the sequence length, allowing UniCSI to process CSI measurements with arbitrary numbers of subcarriers using the same model parameters.

\paragraph{Spectral Aggregator}
After RF tokenization, each antenna link is represented as a sequence of subcarrier tokens, $\mathbf{Z}\in\mathbb{R}^{K\times P\times d}$, where $K$ varies across Wi-Fi devices and communication parameters. To decouple the downstream encoder from this variable spectral length, we introduce a query-based spectral aggregator that compresses each spectrum into a fixed-size representation. Specifically, a set of $Q$ learnable query tokens $\{q_j\}_{j=1}^{Q}$ cross-attend to the $K$ subcarrier tokens at each temporal window, yielding a compact latent representation of size $Q\times P$.

To give queries and subcarrier tokens a shared frequency coordinate, we incorporate a physics-informed RF positional encoding. For a measurement with bandwidth $B$ and subcarriers indexed $k \in \{0,\dots,K-1\}$, the $k$-th subcarrier is assigned the centered frequency offset
\begin{equation}
f_k = \left(k-\frac{K-1}{2}\right)\frac{B}{K},
\end{equation}
normalized by the half bandwidth as
\begin{equation}
\tilde{f}_k=\frac{f_k}{B/2}\in[-1,1],
\label{eq:frac}
\end{equation}
and projected into the embedding space via a learned affine map,
\begin{equation}
\mathrm{PE}(\tilde{f}_k)=W\tilde{f}_k+b,
\qquad W\in\mathbb{R}^{d\times1},\ b\in\mathbb{R}^{d}.
\label{eq:pe}
\end{equation}
Unlike discrete positional embeddings, whose indices are not comparable across bandwidths or channel configurations, this formulation requires neither a lookup table nor a predefined maximum subcarrier count: subcarriers occupying the same relative location within different bandwidths receive consistent positional codes, independent of the subcarrier spacing.

The RF positional encoding is added to both the key/value subcarrier tokens and the query tokens.
Each query is associated with a fixed fractional anchor $\tilde{a}_j\in[-1,1]$ and augmented with $\mathrm{PE}(\tilde{a}_j)$, enabling queries and subcarrier tokens to interact in a shared normalized frequency coordinate system.
Cross-attention is interleaved with Transformer blocks for $R$ refinement steps: at each step, the queries attend to the spectrum, are processed by a Transformer group, and then re-attend to the subcarrier tokens to progressively refine the compressed representation.

The resulting $Q\times P$ latent is further augmented with a learned temporal-window encoding and a band embedding that indicates the operating frequency band (2.4 or 5\,GHz). The tokens are then flattened, optionally prepended with a class token for supervised learning, and processed by a pre-norm Transformer encoder of depth $L$. To support arbitrary MIMO configurations, each transmit--receive link is encoded independently, and the resulting per-link predictions are aggregated at inference time.

\paragraph{Training Objectives}
The UniCSI encoder is agnostic to the learning paradigm: the same architecture is optimized under either supervised multi-task learning or self-supervised pretraining, with only the objective changing.

\textbf{Supervised multi-task learning.} Given $M$ Wi-Fi sensing datasets with heterogeneous tasks, we jointly train the shared encoder $f_{\theta}$ with a lightweight prediction head $g_m$ per dataset, attached to the representation $\mathbf{z}=f_{\theta}(\mathbf{H})$. The objective is a weighted sum of the per-dataset losses, $\mathcal{L}_{\mathrm{sup}}=\sum_{m=1}^{M}\lambda_m\,\mathcal{L}_m(g_m(\mathbf{z}),\mathbf{y}_m)$, where each $\mathcal{L}_m$ is a task-specific loss (cross-entropy for classification) and $\lambda_m$ balances the datasets.

\textbf{Self-supervised pretraining.} To obtain a general-purpose representation, we pretrain with a Joint-Embedding Predictive Architecture (JEPA)~\cite{assran2023}: contiguous temporal-spectral blocks are held out as prediction targets while the remaining tokens form the context, and a lightweight predictor maps the context representation to the target representation. The encoder is trained under a smooth-$L_1$ loss on parameter-free layer-normalized targets, with the target representation produced by an EMA copy of the encoder; details of the masking and normalization are given in the JEPA pretraining experiment.

\section{Experiments}

\subsection{Datasets}
\definecolor{ptshade}{gray}{0.93}
\newcommand{\ptrow}{\rowcolor{ptshade}}
\begin{table*}[t]
    \centering
    \footnotesize
    \setlength{\tabcolsep}{3pt}
    \renewcommand{\arraystretch}{1.0}
    \begin{tabularx}{\textwidth}{X l c r r l l}
    \toprule
    Dataset & Task ($n$) & Rx & Tx & Sub. & Band (GHz) & BW (MHz) \\
    \midrule
    \ptrow Wi-Fi Sensing Contest~\citep{han2024} & Presence ($n{=}3$) & 2 & 2 & 248,\,250 & 5 & 80,\,160 \\
    \ptrow Person-in-Wi-Fi 3D~\citep{yan2024} & 3D Pose ($n{=}4$) & $3{\times}3$ & 1 & 30 & 5 & 20 \\    \ptrow XRFV2~\citep{lan2025} & HAR ($n{=}30$) & $3{\times}3$ & 1 & 30 & 5 & 20 \\
    \ptrow Wi-MIR~\citep{islam2024} & HAR ($n{=}17$) & 3 & 3 & 30 & 5 & 20 \\
    \ptrow Behavior Auth.~\citep{shi2017} & User ID ($n{=}12$) & 3 & 1 & 30 & 5 & 20 \\
    \ptrow NTU-Fi Human ID~\citep{yang2022} & User ID ($n{=}14$) & 3 & 1 & 114 & 5 & 40 \\
    \ptrow MM-Fi$^{\ddagger}$~\citep{yang2023b} & HAR ($n{=}27$) & 3 & 1 & 114 & 5 & 40 \\
    \ptrow WiAR~\citep{guo2019} & HAR ($n{=}16$) & 3 & 1 & 30 & 5 & 20 \\
    \ptrow ARIL~\citep{wang2019} & HAR / Loc. ($n{=}6$) & 1 & 1 & 52 & 2.4 & 20 \\
    \ptrow CSIDA$^{\ddagger}$~\citep{zhang2022} & Gesture ($n{=}6$) & 3 & 1 & 114 & 5 & 40 \\
    \ptrow OPERAnet$^{\ddagger}$~\citep{bocus2021} & HAR ($n{=}7$) & 3 & 3 & 30 & 5 & 40 \\
    \ptrow HAR-ORT$^{\ddagger}$~\citep{bellizzi2024} & HAR ($n{=}5$) & 1 & 1 & 242 & 5 & 80 \\
    \ptrow EHUNAM~\citep{dearmas2025} & HAR / Env.\ ID ($n{=}6$) & 1 & 1 & 56--241 & 2.4,\,5 & 20,\,80 \\
    \ptrow CSI-BFI-HAR~\citep{haque2025} & HAR ($n{=}21$) & 3 & 1 & 242 & 5 & 80 \\
    \ptrow OctoNet~\citep{yuan2026} & HAR ($n{=}64$) & $4{\times}2$ & 1 & 114 & 5 & 40 \\
    \ptrow UT-HAR~\citep{yangSenseFiLibraryBenchmark2023a} & HAR ($n{=}7$) & 3 & 1 & 30 & 5 & 20 \\
    \ptrow CSI-Bench$^{\dagger\S}$~\citep{zhu2025} & Fall/HAR/ID/Prox. & 2--4 & 1--2 & 14--58 & 2.4,\,5 & 20--40 \\
    \midrule
    Exposing the CSI~\citep{cominelli2023} & HAR ($n{=}12$) & $3{\times}4$ & 1 & 2048 & 5 & 160 \\    XRF55~\citep{wang2024} & HAR ($n{=}55$) & $3{\times}3$ & 1 & 30 & 5 & 20 \\
    Widar 3.0~\citep{zheng2019} & Gesture ($n{=}22$) & $6{\times}3$ & 1 & 30 & 5 & 20 \\
    WiMANS~\citep{huang2025} & HAR ($n{=}9$) & 3 & 3 & 30 & 2.4,\,5 & 20 \\
    FallDar~\citep{yang2023a} & Fall ($n{=}2$) & 3 & 1 & 30 & 5 & 20 \\
    GaitID~\citep{zhang2020} & User ID ($n{=}11$) & $6{\times}3$ & 1 & 30 & 5 & 20 \\
    NTU-Fi HAR~\citep{yang2022} & HAR ($n{=}6$) & 3 & 1 & 114 & 5 & 40 \\
    SHARP~\citep{meneghello2023} & HAR ($n{=}8$) & 4 & 1 & 242 & 5 & 80 \\
    \bottomrule
    \end{tabularx}
    \caption{CSI datasets for pretraining and evaluation. Shaded rows form the self-supervised pretraining corpus; unshaded rows are held out for downstream evaluation only. $n$ is the number of class labels, Rx/Tx the antenna counts, Sub.\ the subcarrier count, Band the carrier band and BW the channel bandwidth; $D{\times}A$ denotes $D$ receiver devices of $A$ antennas each. Footnote marks are explained in the text.}
    \label{tab:datasets}
\end{table*}

We gathered as many public CSI datasets as we could in order to test the generalization capability of our approach. The resulting collection spans the breadth of human-sensing tasks, including human activity recognition (HAR), pose estimation, gesture recognition, localization, authentication, and fall detection, so that any model evaluated on it must contend not only with configuration heterogeneity but with a diversity of downstream objectives.
We applied two inclusion criteria in dataset selection. First, we required a minimum level of annotation, namely that each dataset reports its carrier frequency, bandwidth, MIMO dimension, and sampling rate. Second, we required the CSI to be free of preprocessing artifacts. The datasets that remain (listed in Table~\ref{tab:datasets}) still differ substantially in their MIMO configurations, ranging from $1{\times}1$ to $6{\times}3$, with some setups combining several receiver devices for simultaneous recording. Four of the pretraining datasets ship with a designated cross-domain test domain, which we exclude from the corpus and reserve for evaluation (marked $^{\ddagger}$: CSIDA scene~1, HAR-ORT subject~10, MM-Fi scene~3, OPERAnet room~2); CSI-Bench contributes all subsets except HAR ($^{\S}$) and aggregates four tasks with $n{=}2/5/7/4$ ($^{\dagger}$). The full collection totals 530.2 hours of recordings, of which the assembled pretraining corpus comprises 425.6~hours over 391{,}277 measurements after holdouts. To harmonize acquisition settings, we resampled every recording to a common 100~Hz rate. We then segmented each recording into non-overlapping windows of at most 5~seconds, retaining shorter recordings as single samples. Finally, as chipsets can encode CSI with a different bit depth, we normalize each segment to zero mean and unit variance, rendering amplitudes comparable across devices before they reach the encoder.

\subsection{Implementation Details}
All models were trained on a single accelerator with roughly $140$\,GB of memory, hosted on an x86 Linux machine with $256$\,GB of system RAM. Pretraining took $14.75$ hours on a single GPU, using approximately $41$\,GiB of memory.
The detailed hyperparameter settings are summarized in Table~\ref{tab:hyperparameters}.
We report either classification accuracy or macro-F1 depending on the benchmark. Macro-F1 is reported for imbalanced datasets to better reflect per-class performance.

\begin{table}[t]
  \centering
    \caption{Training hyperparameters. All runs use AdamW, cosine decay to $0$, gradient clip $1.0$, and bf16.}
  \label{tab:hyperparameters}
  \footnotesize
  \setlength{\tabcolsep}{4pt}
  \renewcommand{\arraystretch}{1.15}
  \begin{tabularx}{\linewidth}{@{}l CCCC@{}}
    \toprule
    & Multi-task & JEPA & Probe / $k$-NN & Fine-tune \\
    \midrule
    Encoder      & scratch & EMA & frozen & fine-tuned \\
    Head         & linear & predictor & BN + lin. & BN + lin. \\
    Objective    & CE & smooth-$L_1$ & CE / cosine & CE \\
    LR      & $3{\times}10^{-4}$ & $1{\times}10^{-4}$ & $1{\times}10^{-3}$ & $3{\times}10^{-4}$ \\
    WD & $5{\times}10^{-2}$ & $5{\times}10^{-2}$ & $1{\times}10^{-5}$ & $5{\times}10^{-2}$ \\
    Budget       & 50 epochs & 50\,k steps & 200 epochs & 50 epochs \\
    Batch size   & 128 & 1024  & 64 & 64 \\
    \bottomrule
  \end{tabularx}
\end{table}

\subsection{Multi-task Supervised Training}
\label{sec:multi-task}

\begin{figure}
    \centering
    \includegraphics[width=\linewidth]{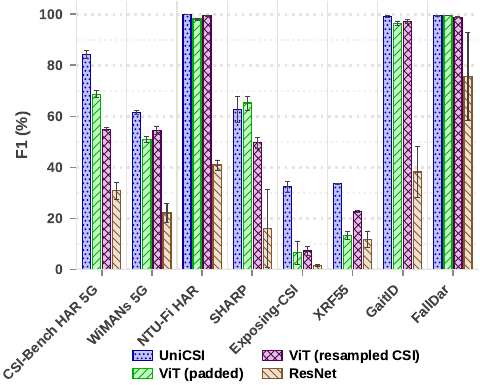}
    \caption{Fused F1 of a single encoder trained jointly across eight heterogeneous CSI datasets. Bars denote the mean and whiskers the standard deviation over three seeds. All datasets except NTU-Fi HAR (in-distribution control) use a cross-environment or cross-subject protocol.}
    \label{fig:e1_results}
\end{figure}

To assess whether a single encoder can benefit from jointly observing heterogeneous CSI configurations, we trained one encoder on eight CSI datasets for HAR spanning four Wi-Fi bandwidths.

We compare UniCSI against baselines chosen to instantiate the shared design axis of prior CSI foundation models, namely a fixed input grid with index-based positional embeddings \citep{jiang2025,zhu2026,luo2026,kim2026}. Rather than reproduce each method verbatim, which would entangle the architectural question with differences in corpora, preprocessing, and training recipes, we abstract their common ingestion strategies and train each under an identical recipe, corpus, and parameter budget. The \emph{Padded ViT} zero-pads CSI to the largest subcarrier dimension before a standard ViT, instantiating the fixed-grid tokenization underlying AM-FM and CSI-JEPA, whereas the \emph{Resampled ViT} interpolates all inputs to a fixed resolution, instantiating the harmonize-then-train strategy of \citet{jiang2025}. A shared \emph{ResNet} serves as a shape-agnostic convolutional control. This design attributes any performance difference to the ingestion mechanism itself rather than to confounded training choices.

Our model achieves the highest macro-F1 across the eight test sets (71.6\%), outperforming the strongest baseline (padded ViT, 62.4\%) by 9.2~points. This advantage is concentrated in the hardest cross-domain regimes. On Exposing-CSI, a 12-way task under an unseen environment, the padded and resampled ViTs reach only 6.7\% and 7.3\% and the ResNet collapses to 1.7\%, leaving every baseline at or below the chance floor, whereas our model attains 32.5\%, marking the difference between transferring to the unseen environment and failing to generalize at all. XRF55 shows the same pattern, with our model at 33.6\% against 22.9\% for the strongest baseline.
On the well-conditioned datasets our model does not sacrifice performance, saturating NTU-Fi HAR (100.0\%), GaitID (99.2\%), and FallDar (99.4\%). The only datasets on which it does not lead are SHARP (62.6\% against the padded ViT's 65.2\%), where the two are within overlapping seed spread, and FallDar, where the padded ViT edges ahead by 0.2~points with both models saturated. The ResNet baseline, which shares a single backbone across shapes without frequency information, is the weakest throughout and the least stable across seeds (average 29.7\%), underscoring that a shape-agnostic convolutional backbone alone does not solve heterogeneous ingestion.

\subsection{JEPA Pretraining}
We pretrain with a joint-embedding predictive architecture (JEPA)~\cite{assran2023} on the unlabeled corpus, drawing measurements with a bandwidth-stratified sampler that enforces a $50/25/25$ mix of $20$, $40$, and ${\geq}\,80$\,MHz links, with the $160$\,MHz data pooled into the widest stratum. For each batch we mask a single contiguous block covering $80\%$ of the encoder's latent token grid ($Q=16$ frequencies by $P=20$ time patches), leaving a thin border of visible tokens as context. The mask is enforced in input space, by zeroing the corresponding subcarrier-by-time region of the raw CSI within each clip's non-padded extent.

The prediction target is produced by an EMA target encoder whose momentum is annealed linearly from $0.996$ to $0.9995$, encoding the unmasked input. A narrow predictor ($4$ layers, width $192$) then maps the visible context tokens, together with a learned mask token at each masked grid position, to the target representations at those positions. We regress onto parameter-free layer-normalized targets under a smooth-$L_1$ loss, since normalizing the targets removes the scale-collapse shortcut that the encoder's affine output norm would otherwise leave open, by which the predictor could trivially minimize the loss by driving target magnitudes toward zero rather than learning to predict their structure.

\subsubsection{Transferability of Pretrained Representations}
\begin{figure*}[t]
  \centering
  \includegraphics[width=.85\linewidth]{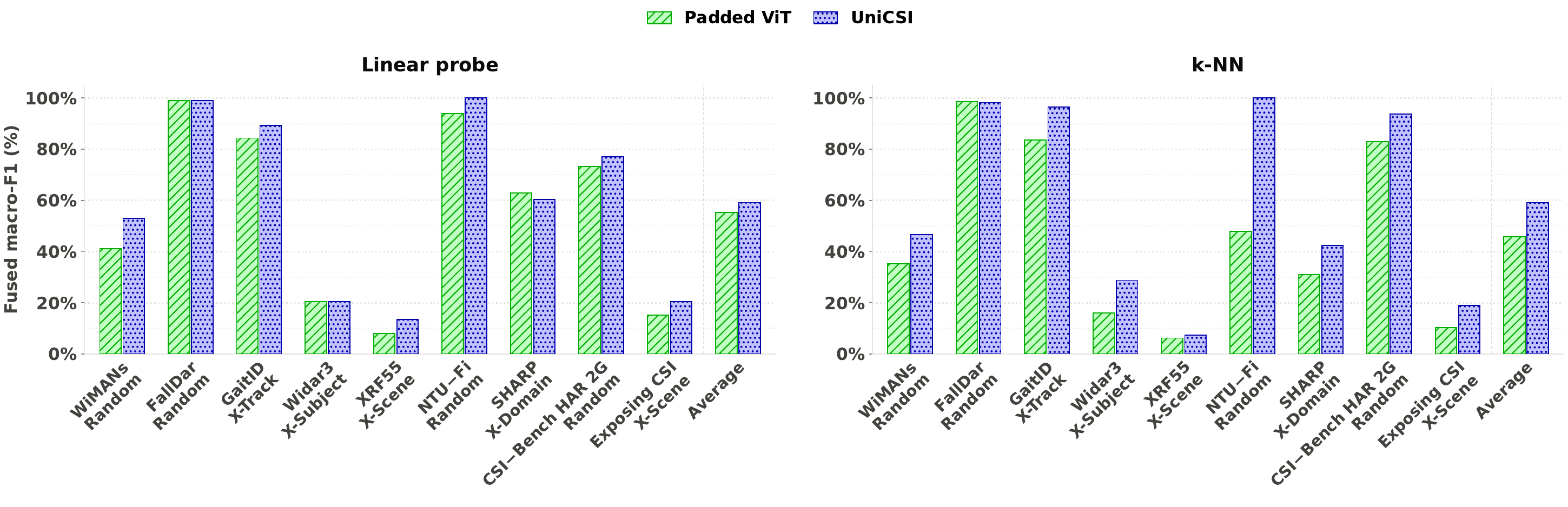}
  \caption{Linear probe and $k$-NN evaluation of Padded ViT and UniCSI on eight representative dataset splits. Bars report fused macro-F1; Avg is the mean over the selected splits.}
  \label{fig:probe-knn-f1}
\end{figure*}

To assess the quality and transferability of the learned representations, we evaluate the pretrained encoder under two frozen-encoder protocols, $k$-NN classification and linear probing, in which the encoder is held fixed and only a lightweight predictor is fit (Figure~\ref{fig:probe-knn-f1}). For $k$-NN, we select the best $k$ based on validation set performance and then apply the best $k$ to the test set. Both protocols measure the discriminative power of the representation itself, without updating the backbone. Across most downstream datasets and under both protocols, UniCSI outperforms the pretrained Padded ViT, the strongest supervised baseline. This baseline is itself an informative comparator, since a JEPA-pretrained Padded ViT is CSI-JEPA-like in all but one respect, sharing its fixed patch grid, index-based positional embeddings, and joint-embedding predictive objective, and trained on the same corpus as UniCSI, so that it differs from \citet{luo2026} only in using uniform rather than variation-aware masking. The gains are most pronounced on the harder cross-environment benchmarks, namely WiMANS, SHARP, and Exposing CSI, while performance remains comparable on easier datasets such as FallDar. That the improvement is larger under $k$-NN is telling, as it suggests UniCSI learns representations that are more transferable and more semantically structured even without any downstream adaptation.

We further evaluate parameter adaptation by allowing the pretrained encoder to be optimized using full fine-tuning. The results are reported in Table~\ref{tab:full_ft}.
We compare three settings: supervised training from scratch using UniCSI, full fine-tuning of a pretrained Padded ViT, and full fine-tuning of a pretrained UniCSI. As shown in Table~\ref{tab:full_ft}, pretrained UniCSI achieves the highest average performance (71.58\%), outperforming both training UniCSI from scratch (67.97\%) and the pretrained Padded ViT (60.68\%). The improvement over training from scratch demonstrates that large-scale pretraining provides a stronger initialization for downstream adaptation, while the substantial margin over Padded ViT indicates that preserving the native spectral structure is more effective than relying on zero-padding to accommodate heterogeneous CSI.

Compared with the baseline pretrained model, UniCSI achieves stronger downstream performance on the majority of benchmarks under both frozen and trainable adaptation protocols, with the largest margins on cross-environment splits.

\begin{table}[t]
\centering
\caption{Comparison between supervised training from scratch with UniCSI, full fine-tuning (FT) from the JEPA-pretrained Padded ViT, and full FT from the JEPA-pretrained UniCSI on unseen downstream datasets (Metric: Macro F1 score).}
\label{tab:full_ft}
\footnotesize
\setlength{\tabcolsep}{4pt}
    \begin{tabularx}{\columnwidth}{X l c c c}
        \toprule
        Dataset & Split & \makecell{Scratch\\(Ours)} & \makecell{Full FT\\(PaddedViT)} & \makecell{Full FT\\(Ours)} \\
        \midrule
        WiMANS (5\,GHz) & Cross-Env & \underline{55.69\%} & 36.96\% & \textbf{55.83\%} \\
        XRF55 & Cross-Env & \underline{39.53\%} & 15.5\% & \textbf{39.78\%} \\
        SHARP & Cross-Env & 58.50\% & \textbf{75.71\%} & \underline{71.06\%} \\
        Exposing-CSI & Cross-Env & \textbf{47.56\%} & 7.6\% & \underline{45.16\%} \\
        FallDar & Random & 96.78\% & \underline{98.61\%} & \textbf{98.95\%} \\
        CSI-Bench (2.4\,GHz) & Random & 80.19\% & \textbf{91.16\%} & \underline{90.67\%} \\
        NTU-Fi HAR & Random & 97.52\% & \underline{99.23\%} & \textbf{99.61\%} \\
        \midrule
        Average & &\underline{67.97\%} & 60.68\%& \textbf{71.58\%} \\
        \bottomrule
    \end{tabularx}
\end{table}

\subsubsection{Sample Efficiency}

We next evaluate the sample efficiency of UniCSI under limited supervision. Starting from the pretrained checkpoint or random initialization, we fine-tune all model parameters using only 5, 10, or 50 labeled samples per class on eight downstream datasets spanning in-domain, random, and cross-domain evaluation protocols.

\begin{figure}[t]
  \centering
  \includegraphics[width=\linewidth]{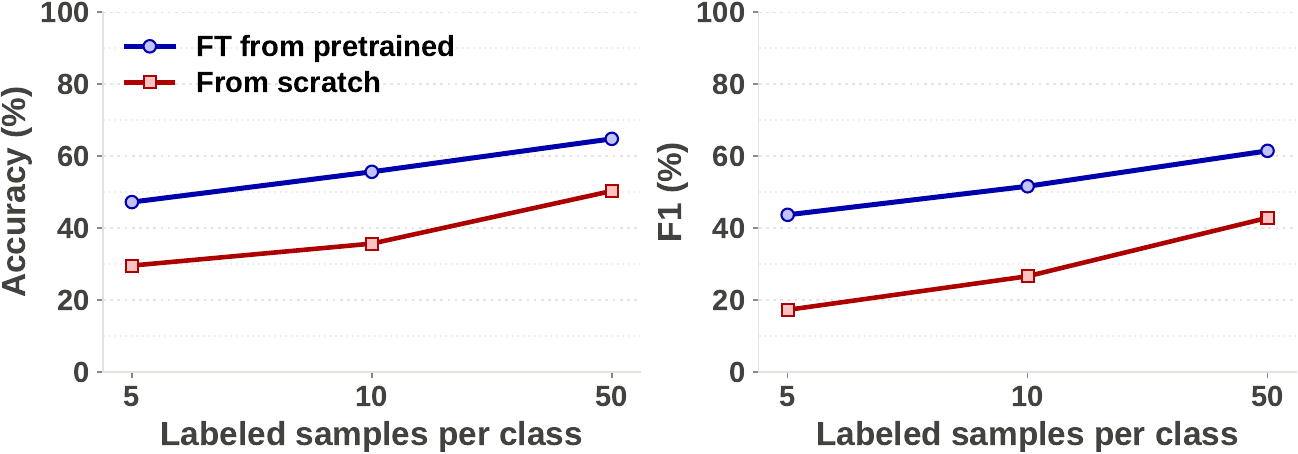}
    \caption{Few-shot downstream fine-tuning from pretrained versus random initialization, averaged over eight dataset--split pairs.}
  \label{fig:fewshot-ft-lines}
\end{figure}

Figure~\ref{fig:fewshot-ft-lines} summarizes the average performance across all datasets. Pretraining consistently improves downstream performance at every label budget, with the largest gains in the most label-scarce setting. With only 5 labeled samples per class, pretrained UniCSI increases the average fused accuracy from 29.6\% to 47.2\% and the macro-F1 from 17.3\% to 43.7\%. The advantage remains substantial at 10-shot and 50-shot, demonstrating that the learned representation can be adapted effectively with very limited supervision. These results indicate that UniCSI is highly sample-efficient, substantially reducing the amount of labeled data required for downstream deployment.

\subsection{Ablation}
We isolate the contribution of the two main components by removing them one at a time from the full model, under the same training scheme as multi-task supervised training. Both variants share the same training mixture, schedule, optimizer, and parameter budget, and each is run with three seeds; we report the average accuracy across the eight datasets of the multi-task scheme, since several components produce differences comparable to seed noise.
Both components contribute, but to very different degrees. Removing the RF tokenizer produces the largest drop, from 73.3\% to 63.5\%: replacing per-subcarrier tokens with fixed two-dimensional patches forces samples onto a common grid and discards the per-subcarrier temporal processing, which suggests that the tokenizer is the component the rest of the architecture is built on. Removing spectral aggregation costs only 1.6 points (73.3\% to 71.7\%), which may appear to make the component optional; however, the average conceals its role, as the aggregation is what recovers the accuracy that per-subcarrier tokenization sacrifices on low-subcarrier devices, and it is also what fixes the latent size for any subcarrier count, the property the JEPA pretraining depends on.

\section{Discussion}
Across both training regimes, our results support a single claim: handling CSI heterogeneity in the architecture, rather than normalizing it away beforehand, is what enables transfer across devices and environments. Under supervised joint training UniCSI is the only model that transfers to unseen environments where the fixed-grid baselines collapse toward chance, and under self-supervised pretraining the same encoder yields representations that transfer under frozen probing and initialize few-shot fine-tuning far better than training from scratch. The ablation locates these gains in the two components that make the architecture shape-agnostic, the per-subcarrier tokenizer and the spectral aggregator, rather than in scale or data alone, suggesting that native heterogeneous ingestion is not merely a convenience but a property that materially improves cross-domain generalization.

We kept the positional encoding deliberately simple, and see richer frequency representations as the most promising follow-up. Encoding absolute carrier frequency rather than band-relative position is a natural next step.

We compare against abstractions of prior ingestion strategies rather than published checkpoints, since existing models are bound to their preprocessing pipelines and cannot ingest our corpus without the normalization our study removes. Head-to-head evaluation on shared benchmarks under each method's native preprocessing nonetheless remains valuable future work.

\section{Conclusion}
We presented \textbf{UniCSI}, a unified foundation architecture for heterogeneous Wi-Fi sensing that natively accommodates diverse CSI configurations without lossy preprocessing. By combining a physics-informed RF tokenizer with a spectral aggregator, UniCSI enables a single model to process CSI collected under varying subcarrier counts, bandwidths, and carrier bands. Extensive experiments under both supervised and self-supervised settings show that handling heterogeneity in the architecture, rather than in preprocessing, substantially improves cross-device and cross-environment generalization, with the largest gains precisely where fixed-grid approaches collapse.

\bibliography{aaai2027}

\end{document}